\documentclass[letterpaper,11pt]{article}

\usepackage[utf8]{inputenc} 
\usepackage[T1]{fontenc}    
\usepackage{hyperref}       
\usepackage{url}            
\usepackage{booktabs}       
\usepackage{amsfonts}       
\usepackage{nicefrac}       
\usepackage{microtype}      
\usepackage{xcolor}         
\usepackage{graphicx}
\usepackage[singlelinecheck=false,center]{caption} 
\usepackage{comment}
\usepackage{multirow}
\usepackage{amssymb,amsmath,amsthm}
\usepackage{wrapfig}
\usepackage{adjustbox}
\usepackage{tabularx}
\usepackage{fullpage}
\usepackage[numbers]{natbib}
\newtheorem{theorem}{Theorem}
\let\vec\mathbf
\title{BayesFormer: Transformer with Uncertainty Estimation}

\newcommand{\KL}{\mathtt{KL}}
\newcommand{\query}{\mathtt{query}}
\newcommand{\key}{\mathtt{key}}
\newcommand{\val}{\mathtt{val}}
\newcommand{\inp}{\mathtt{input}}
\newcommand{\position}{\mathtt{pos}}
\newcommand{\attn}{\mathtt{attn}}
\newcommand{\mlp}{\mathtt{mlp}}
\newcommand{\mha}{\mathtt{mha}}
\newcommand{\base}{\mathtt{base}}
\newcommand{\LN}{\mathtt{LN}}
\usepackage{array}
\newcolumntype{P}[1]{>{\centering\arraybackslash}p{#1}}

\usepackage{authblk}

\author[1]{Karthik Abinav Sankararaman\thanks{karthikabinavs@fb.com}}
\author[2]{Sinong Wang\thanks{sinongwang@fb.com}}
\author[2]{Han Fang\thanks{hanfang@fb.com}}
\affil[1]{Meta AI, Austin, TX}
\affil[2]{Meta AI, Seattle, WA}

\begin{document}

\maketitle

\begin{abstract}
Transformer has become ubiquitous due to its dominant performance in various NLP and image processing tasks. However, it lacks understanding of how to generate mathematically grounded uncertainty estimates for transformer architectures. Models equipped with such uncertainty estimates can typically improve predictive performance, make networks robust, avoid over-fitting and used as acquisition function in active learning. In this paper, we introduce BayesFormer, a Transformer model with dropouts designed by Bayesian theory. We proposed a new theoretical framework to extend the approximate variational inference-based dropout to Transformer-based architectures. Through extensive experiments, we validate the proposed architecture in four paradigms and show improvements across the board: language modeling and classification, long-sequence understanding, machine translation and acquisition function for active learning.
\end{abstract}

\section{Introduction}

Transformer-based architectures~\cite{vaswani2017attention} are now the primary state-of-the-art (SOTA) models for a number of tasks in the computer vision~\cite{dosovitskiy2020image}, natural language processing~\cite{vaswani2017attention} and speech recognition~\cite{speechTransformer} domains. They have shown superior capabilities in a multitude of settings including classification, regression, time-series forecasting~\cite{li2019enhancing} and reinforcement learning~\cite{chen2021decision}. For instance, within the natural language processing domain, the notion of \emph{large language models} (LLM) such as BERT~\cite{devlin2018bert} and RoBERTa~\cite{liu2019RoBERTa} have emerged which are pretrained on internet scale data and then used to perform few-shot learning by fine-tuning on target data labels. This paradigm has led to breakthrough results in sentiment analysis, text classification and question answering. Due to this superiority in predictive capabilities, transformers have started to be used as the primary modeling technique in a number of real-world machine learning systems.

\begin{figure}[!htb]
   \begin{minipage}{0.48\textwidth}
     \centering
     \includegraphics[width=\linewidth]{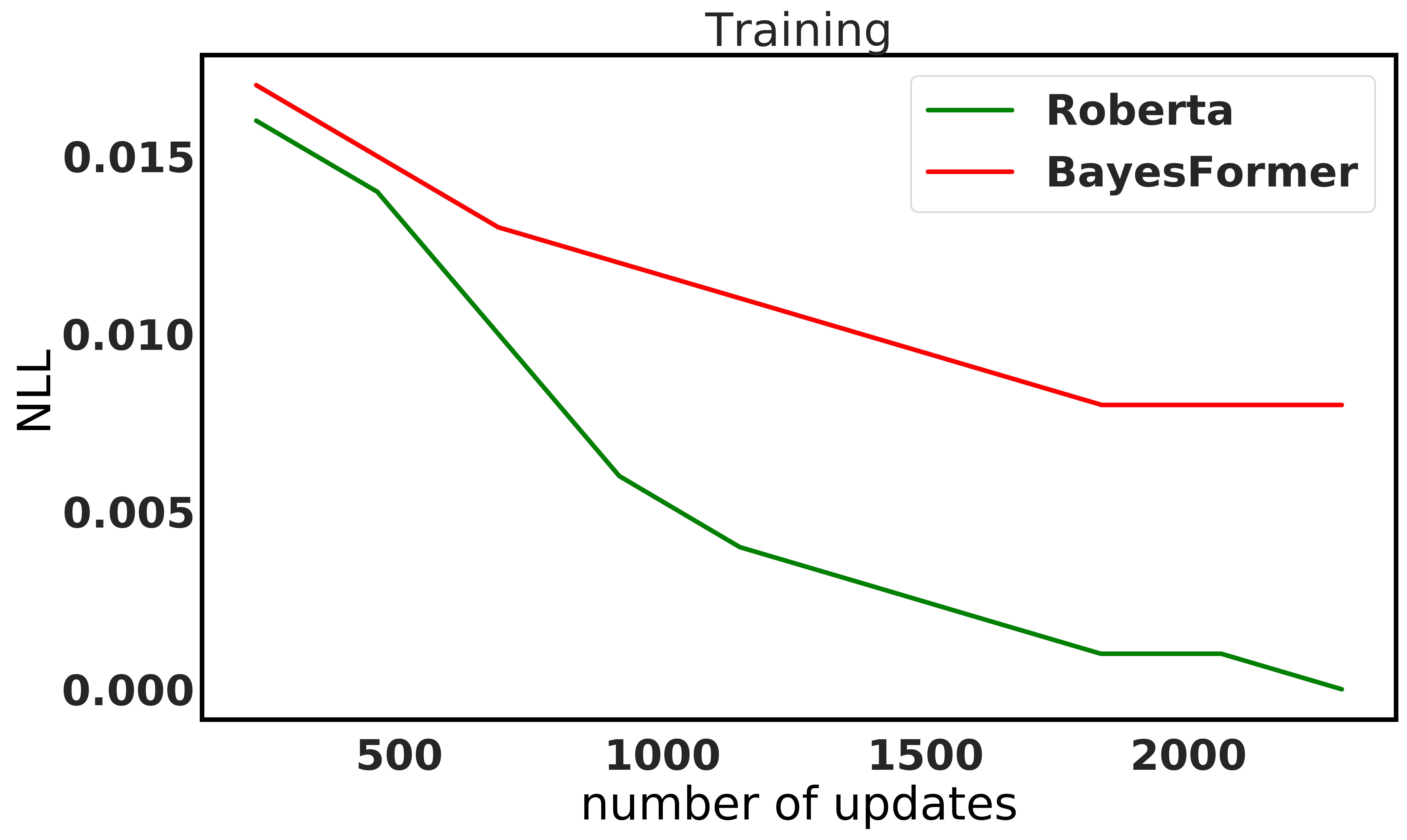}
   \end{minipage}\hfill
   \begin{minipage}{0.48\textwidth}
     \centering
     \includegraphics[width=\linewidth]{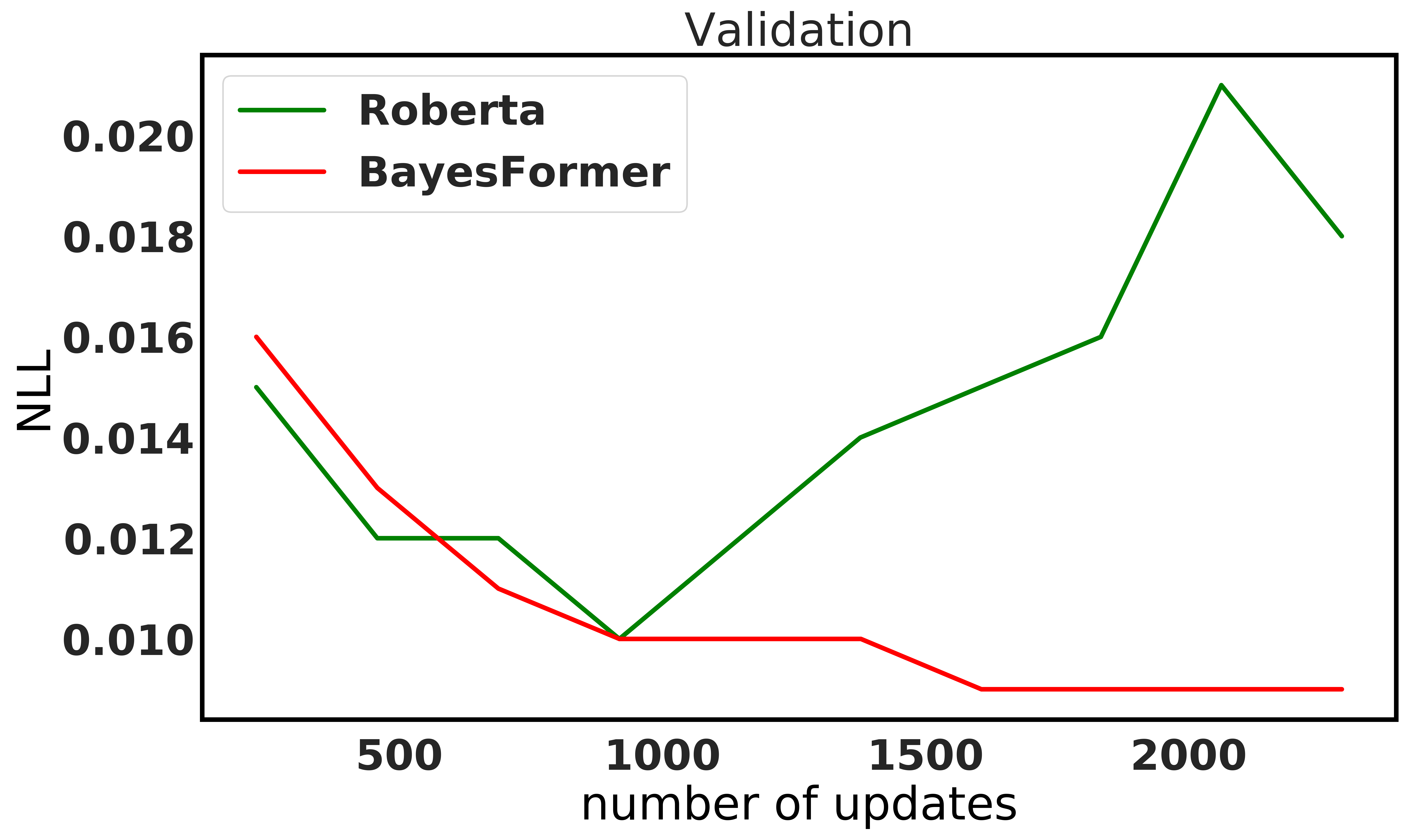}
   \end{minipage}
   \caption{Negative log-likelihood evolution when fine-tuning LLM on MRPC dataset}
   \label{Fig:overfit_nll}
\end{figure}

Despite the progress made on improving the predictive capabilities of transformers, lesser attention has been devoted to understanding the quality of predictions on \emph{individual} data. In particular, when integrating transformers into real-world systems a practitioner is concerned about details beyond \emph{average} performance such as robustness, fairness, over-fitting and confidence of predictions. For example, Figure~\ref{Fig:overfit_nll} depicts a LLM pretrained on 16GB of internet data (similar to BERT) and then finetuned on a paraphrase classification dataset (MRPC~\cite{dolan2005automatically}) for the purpose of predicting whether a given sentence is a paraphrase or not. We see that the commonly used RoBERTa$_\base$ LLM~\cite{liu2019RoBERTa} quickly overfits during finetuning. A primary aspect of understanding the quality of individual predictions is via uncertainty quantification. At its core, uncertainty quantification is concerned with the ability of models to output a notion of \emph{confidence} of a prediction along with the prediction.

Uncertainty quantification for machine learning models is a rich-area of study (\emph{e.g.,} \cite{lakshminarayanan2017simple,blundell2015weight,louizos2016structured}). It is the key tool used to help a system convert a point-prediction to a reliable decision. Some applications include helping balance exploration-exploitation in recommendation systems~\cite{bayesExploration}, as acquisition functions in active learning~\cite{lewis1994sequential}, improving robustness on individual predictions~\cite{hendrycks2019benchmarking}, avoiding over-fitting on unseen datasets~\cite{gal2016theoretically}. Seminal works have shown various methods to derive uncertainty estimates for feed-forward neural networks~\cite{gal2016dropout}, recurrent neural networks such as LSTMs and GRUs~\cite{gal2016theoretically} and Convolutional neural networks~\cite{galThesis}. However, we have limited understanding of these techniques in the context of Transformers. There is only one recent work~\cite{gleave2022uncertainty} that aims to use last-layer ensemble to obtain uncertainty estimates and apply it in the context of active learning.

\emph{In this work, we seek to scientifically understand how to provide uncertainty quantification on individual predictions of transformers that are mathematically grounded.} 

The main contribution of this paper\footnote{Due to space limitation we defer a comprehensive related work section to the Appendix. In the main section, we mention works directly relevant to our paper.} is to use the approximate variational inference lens of dropout~\cite{gal2016dropout} applied to transformer architectures to derive the BayesFormer architecture. BayesFormer can be used to obtain \emph{interpretable} uncertainty estimates on individual predictions and thus, help alleviate many of the issues mentioned above such as overfitting in small datasets. We provide the theory supporting BayesFormer and validate empirically by showing improved \emph{aggregate} performance in a variety of paradigms including large language models' pre-train-then-fine-tune applications, large-range context understanding problems, machine translation and active learning. The key architectural difference in BayesFormer is the application of dropout masks; both where and how they are applied. This is derived using approximate variational inference to compute the posterior distribution over the weights, given a finite dataset and the prior initialization distribution.

\section{Background}

In this section we give the required background on approximate variational inference, Transformers and Bayesian neural networks.

\subsection{Bayesian neural networks and approximate variational inference}

The goal of approximate variational inference in connection to Bayesian neural networks is to characterize the posterior distribution of the network weights given the data and prior initialization. The typical prior distribution placed over the weights is $\mathcal{N}(0, \sigma^2 \vec{I})$ for a small value $\sigma$. Denote $p(\vec{W}~|~\vec{X}, \vec{Y})$ as the posterior distribution of the weights given the dataset $(\vec{X}, \vec{Y})$. Equipped with an exact computation of this probability, one could then perform prediction on a new data point $\vec{x}$ using the formula
\[
    p(\vec{y}~|~\vec{x}) = \int p(\vec{y}~|~\vec{x}, \vec{W})\cdot p(\vec{W}~|~\vec{X}, \vec{Y}) \cdot d \vec{W}
\]
However, the distribution $p(\vec{W}~|~\vec{X}, \vec{Y})$ is extremely complex; we bypass by leveraging approximate variational inference (\emph{e.g.,} \cite{gal2016dropout,gal2016theoretically,graves2011practical,blundell2015weight}), where we approximate the distribution by a surrogate distribution $q(\vec{W})$ that is easier to compute and minimize the evidence lower bound $\KL(q(\vec{W}) || p(\vec{W}~|~\vec{X}, \vec{Y}))$ over all possible surrogates $q(.)$. 

Thus, one needs to solve the following variational optimization problem
\begin{equation}
    \label{eq:ELBO}
    \min_{q} \KL(q(\vec{W}) || p(\vec{W}~|~\vec{X}, \vec{Y}))
\end{equation}

It can be shown that this is equivalent to the following minimization program.

\begin{equation}
    \label{eq:ELBOAlternate}
    \min_{q} - \int q(\vec{W}) \cdot \log [p(\vec{Y}~|~\vec{X}, \vec{W})] \cdot d\vec{W} + \KL(q(\vec{W}) || p(\vec{W}))
\end{equation}

For a finite dataset $\{\vec{x}_i, \vec{y}_i\}_{i=1}^n$ we use the empirical risk minimization framework to equivalently solve the following optimization problem.

\begin{equation}
    \label{eq:ELBOAlternateERM}
    \min_{q} - \sum_{i=1}^n \int q(\vec{W}) \cdot \log [p(\vec{y}_i~|~\vec{x}_i, \vec{W})] \cdot d\vec{W} + \KL(q(\vec{W}) || p(\vec{W}))
\end{equation}

\subsection{Transformer Architecture}
\label{subsec:trasnformer}
    The key component of a transformer is the multi-head attention unit. Each head within this attention unit $a$ is parameterized by three matrices $\vec{W}_{Q, a}$, $\vec{W}_{K, a}$ and $\vec{W}_{V, a}$ corresponding to the \emph{query}, \emph{key} and \emph{value} matrices $\vec{X}_{\query}, \vec{X}_{\key}, \vec{X}_{\val} \in \mathbb{R}^{n \times d}$ respectively. Here $n$ is the length of the input sequence and $d$ is the length of the embedding.
    
    \begin{equation}
        \label{eq:multi-headAttention}
        \vec Z_{a}(\vec X) = \sigma\left(\frac{\vec{X}_{\query} \cdot \vec{W}_{Q, a} \cdot \vec{W}_{K, a}^T \cdot \vec{X}_{\key}^T}{\sqrt{d}} \right) \cdot \vec{W_{V, a}} \cdot \vec{X}_{\val}
    \end{equation}
    
    In the typical usage of the attention unit, the so-called self-attention is used on the input, where $\vec{X}_\attn := \vec{X}_{\query} = \vec{X}_{\key} = \vec{X}_{\val}$.
    
    The Transfomer encoder architecture with self-attention is obtained as follows. Given an input $\vec{x}$ we obtain a sequence of positions $\vec{p}$. We learn an embedding $X_{\inp}$ and $X_{\position}$ by learning the corresponding weights $W_{\inp}$ and $W_{\position}$ and concatenating the outputs followed by a layer-norm to obtain $\vec{X}_{\attn}$. The architecture itself is then a stack of self-attention followed by a MLP block.
    
    Of particular interest to the main results of this paper are the learnable weights $\vec{W}_{\inp}, \vec{W}_{\position}$ the weights in the self-attention units $\vec{W}_{Q, a}, \vec{W}_{K, a}, \vec{W}_{V, a}$ and the weights in the MLP block $\vec{W}_{\mlp}$.

\section{Variational Inference, Dropout and BayesFormer}

In this section, we derive the new dropout procedure for transformers obtained by using approximate variational inference to find an approximate optimal solution $q(.)$ to the Equation~\eqref{eq:ELBO}.

We view \emph{each} learnable parameter in the Transformer model as a random variable. For ease of notation, we merge the bias term (if any) into the corresponding weight matrix and the associated input vector. Additionally, for ease of presentation we consider the encoder only architecture of the Transformer. This can easily be extended to encoder-decoder transformer architectures.


For a given example $(\vec{X}, \vec{y})$, let $\vec{f}_{\vec{y}}$ denote the output of the encoder architecture of the transformer model. To mathematically describe the transformer architecture we need to setup some notations. Let $\ell$ denote the number of multi-head attention layers and let $k$ denote the number of heads within each layer. Let $\vec{X}_\attn$
denote the input to the multi-head attention layers. Recall from subsection~\ref{subsec:trasnformer} $\vec{W}_{\inp}, \vec{W}_{\position}$ denote the weights to learn an input embedding and the position embedding respectively. Thus $\vec{X}_\attn$ can be written as 

\begin{equation}
    \label{eq:TransInput}
    \vec{X}_\attn := \vec{X}_\inp \cdot\vec{W}_{\inp} \odot \vec{X}_\position \cdot \vec{W}_{\position}. 
\end{equation}

For layer $i$ and head $j$, we define $\vec{Z}_{i, j}(\vec{X}_\mha)$ as the self-attention as defined in Equation~\eqref{eq:multi-headAttention}. We combine the outputs of the $k$ heads at layer $i$ and define $\vec{Z}_i := \vec{Z}_{i, 1} \odot \vec{Z}_{i, 2} \odot \ldots \odot \vec{Z}_{i, k}$. Let $\LN(.)$ denote the layer norm function and let $\vec{X}_{\ell, \attn}$ denote the input to layer $\ell$. Here the function $\sigma(.)$ is a pointwise operator applied on each entry of the matrix.

\begin{equation}
    \label{eq:transformerFullOutput}
    \vec{f}_{\vec{y}}(\vec{X}_\attn) = \sigma(\vec{W}_{\mlp, \ell}^T \cdot \LN(\vec{Z}_{\ell}( \sigma(\vec{W}_{\mlp, \ell-1}^T \cdot \LN(\vec{Z}_{\ell-1}(\ldots \vec{X}_\attn)\ldots) + \vec{X}_{\ell, \attn}))
\end{equation}

Let $\vec{W}$ denote the set of random matrices corresponding to the learnable parameters 
$$\{ \{\vec{W}_{\mlp, i}\}_{i=1}^\ell, \{\vec{W}_{Q, i, j}, \vec{W}_{K, i, j}, \vec{W}_{V, i, j}\}_{i=1,j=1}^{i=\ell, j=k}\}$$ with the prior distribution being the standard normal distribution independent across the matrices and entries within the matrix. Let $\vec{f}_{\vec{y}, \vec{W}}(\vec{X})$ denote the corresponding model output on input $\vec{X}$. Let $\mathcal{D} = \{\vec{y}_i, \vec{X}_i\}_{i=1}^n$ denote the finite dataset at hand. Consider the ERM approximation in Equation~\eqref{eq:ELBOAlternateERM}. For each term $i$ in the summation, we consider the integral
\[
     \int q(\vec{W}) \cdot \log [p(\vec{y}_i~|~\vec{f}_{\vec{y}_i, \vec{W}}(\vec{X}_i))] \cdot d\vec{W}
\]

Similar to \cite{gal2016dropout}, we can approximate the above integral by noting that it computes the expectation of the quantity $\log [p(\vec{y}_i~|~\vec{f}_{\vec{y}_i, \vec{W}}(\vec{X}_i))]$ over $\vec{W}$ that follows the distribution $q(\vec{W})$. Thus, we can obtain a unbiased estimate $\widehat{\vec{W}}_i \sim q(\vec{W})$ and approximate the integral by,
\[
     \int q(\vec{W}) \cdot \log [p(\vec{y}_i~|~\vec{f}_{\vec{y}_i, \vec{W}}(\vec{X}_i))] \cdot d\vec{W} \approx \log [p(\vec{y}_i~|~\vec{f}_{\vec{y}_i, \widehat{\vec{W}}_i}(\vec{X}_i))]
\]

Computing $n$ i.i.d. samples $\widehat{\vec{W}}_1, \widehat{\vec{W}}_2, \ldots, \widehat{\vec{W}}_n \sim q(\vec{W})$ and plugging it back into Equation~\eqref{eq:ELBOAlternateERM} we get the optimization function to be approximately,

\begin{equation}
    \label{eq:MCApprox}
     \min_{q} - \sum_{i=1}^n \log [p(\vec{y}_i~|~\vec{f}_{\vec{y}_i, \widehat{\vec{W}}_i}(\vec{X}_i))] + \KL(q(\vec{W}) || p(\vec{W}))
\end{equation}

\begin{figure}[!htb]
   \begin{minipage}{0.48\textwidth}
     \centering
     \includegraphics[width=.7\linewidth]{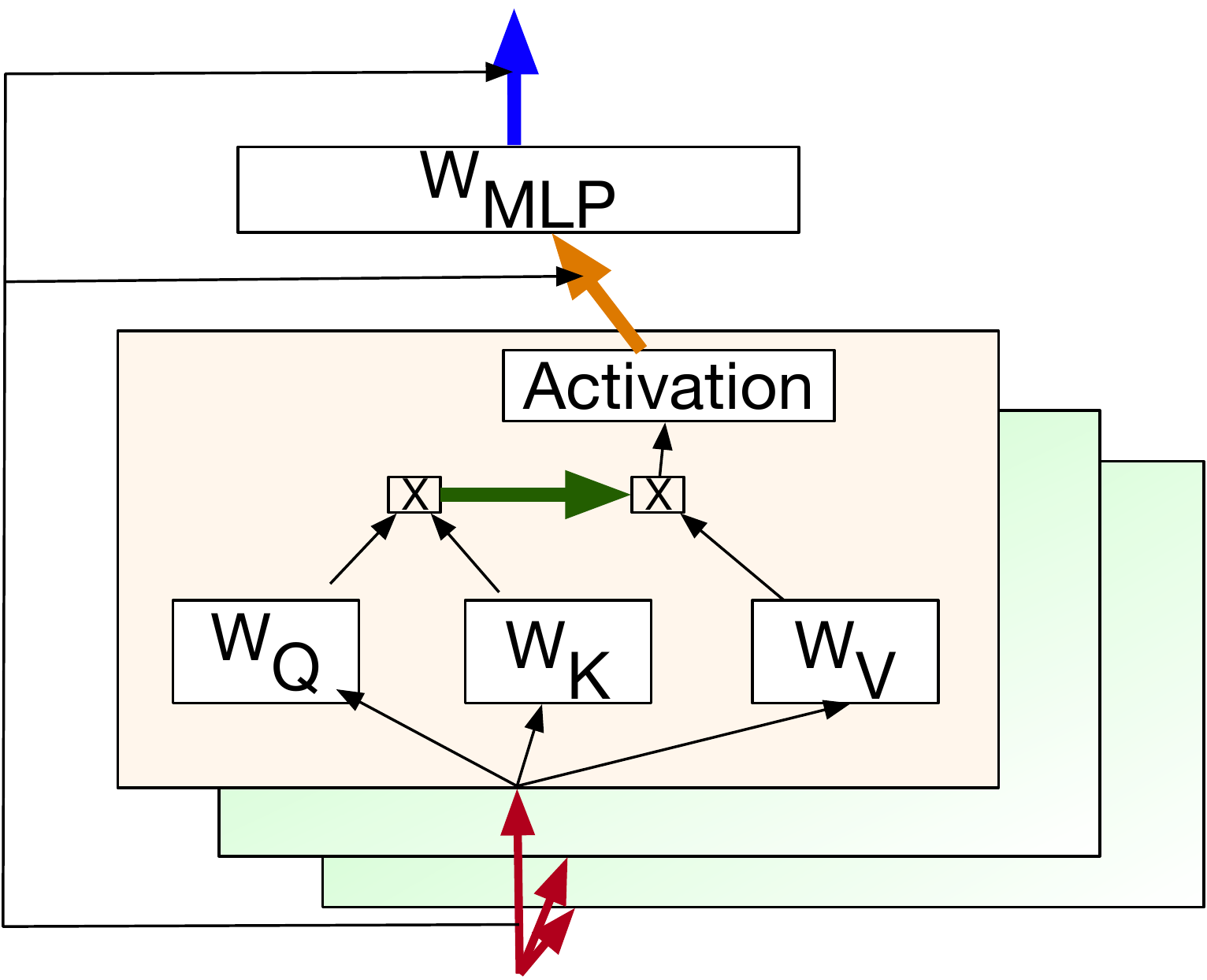}
   \end{minipage}\hfill
   \begin{minipage}{0.48\textwidth}
     \centering
     \includegraphics[width=.7\linewidth]{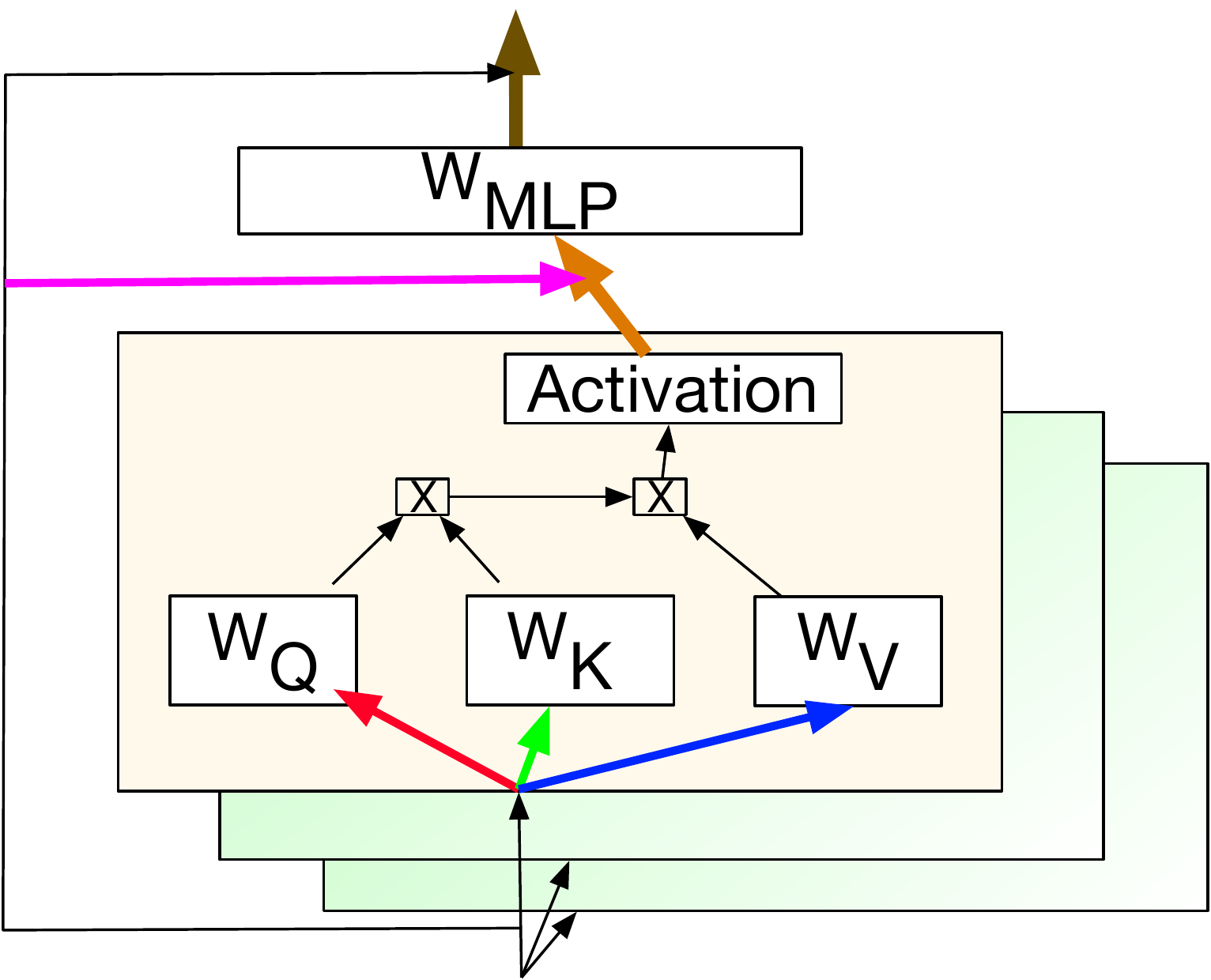}
   \end{minipage}
   \caption{(left) Dropout application in the standard transformer architecture (right) Proposed dropout application derived from approximate variational inference. Same colored arrows indicate the same random outcomes while different colors indicate independent random variables.}
   \label{Fig:BayesFormer_summary}
\end{figure}

To solve the optimization problem in equation~\eqref{eq:MCApprox}, we need to provide two details: 
(a) to solve the minimization problem over a reasonable class of functions $q(.)$. (b) for any given function $q(.)$, the ability to compute an unbiased estimate $\widehat{\vec{W}}$.

\subsection{Variational distribution $q(.)$}
Following the prior works on ANN \cite{gal2016dropout} and RNN \cite{gal2016theoretically}, we define the family of variational distributions $q(.)$ for each matrix in $\vec{W}_{*}$ parametrized by $\vec{M}_*$. For ease of notation, we denote this family by $q_{\vec{M}}(.)$. Let $\vec{W}_{*, i}, \vec{M}_{*, i}$ refer to any given row $i$ in matrices $\vec{W}_{*}$ and $\vec{M}_*$ respectively. The distribution $q_{\vec{M}}(\vec{W})$ is defined as follows. For each $\vec{W}_{\ast} \in \vec{W}$ can be defined as,
\begin{equation}
    \label{eq:variationalDistribution}
   \vec{W}_{{\ast}, i} \sim p \mathcal{N}(0, \sigma^2 \vec{I}) + (1-p) \mathcal{N}(\vec{M}_{*, i}, \sigma^2 \vec{I}).
\end{equation}
In equation~\eqref{eq:variationalDistribution}, the quantity $p$ represents the apriori defined probability and $\sigma$ is the quantity used in the prior initialization of the weights.

Plugging this variational distribution $q_{\vec{M}}(.)$ back in the optimization problem in equation~\eqref{eq:MCApprox}, we get

\begin{equation}
    \label{eq:MCApproxParameterized}
     \min_{q_\vec{M}} - \sum_{i=1}^n \log [p(\vec{y}_i~|~\vec{f}_{\vec{y}_i, \widehat{\vec{W}}_i}(\vec{X}_i))] + \KL(q_{\vec{M}}(\vec{W}) || p(\vec{W}))
\end{equation}

As shown in \cite{gal2016dropout}, when the prior distribution $p(\vec{W})$ are all initialized using the standard normal distribution, we obtain the standard loss maximum likelihood loss function with $\ell_2$ regularization over the weight matrices. In particular, the first term minimizes the negative log-likelihood, while the KL(..) term for the above variational distribution family $q(.)$ (approximately) reduces to minimizing the square of the $\ell_2$-norm of $\vec{M}$ when the prior distribution $p(\vec{W})$ is the standard normal distribution. Thus, for this definition of $q(.)$, the optimal solution to the approximate variational inference evidence lower bound is \emph{equivalent} to the standard trasnformer training objective.

\subsection{BayesFormer and obainting unbiased estimate $\widehat{\vec{W}}$}

Given the above variational distribution $q(.)$ we now provide details on obtaining an unbiased estimate $\widehat{\vec{W}}$. To do this, we introduce the BayesFormer archietecture which modifies the application of dropout to the standard transformer architecture. We then obtain the following theorem.

\begin{theorem}
    \label{thm:mainTheorem}
    For any input $\vec{X}$, we have the following properties of the BayesFormer architecture.
    \begin{enumerate}
        \item Computing the model output $\vec{f}_{\vec{y}, \widehat{\vec{W}}}(\vec{X})$ is equivalent to performing a forward pass on the input $\vec{X}$ and applying the dropout with probability $p$ in the BayesFormer architecture.
        \item  The posterior $p(\vec{y}~|~\vec{X}) \approx \frac{1}{T} \sum_{t=1}^T \vec{f}_{\vec{y}, \widehat{\vec{W}}_t}(\vec{X})$ where $\vec{f}_{\vec{y}, \widehat{\vec{W}}_t}(\vec{X})$ is the output of the $t^{th}$ forward pass through the BayesFormer architecture. Moreover, the predictive uncertainty over the prediction $p(\vec{y}^*~|~\vec{X}^*)$ can be approximated by the empirical lower and upper confidence intervals\footnote{\emph{e.g.,} by performing bootstrapping.} of the $T$ forward passes\footnote{This is the so-called MCDropout procedure from \cite{gal2016dropout}.}.
    \end{enumerate}
\end{theorem}

In the rest of this subsection, we prove Theorem~\ref{thm:mainTheorem} by describing the BayesFormer architecture. From the definition of the variational distribution in equation~\eqref{eq:MCApproxParameterized} we have that each row of each of the weight matrix is \emph{zeroed out} with probability $p$. Motivated by this, in BayesFormer we independently apply dropout mask to the vectors $\vec{X}_\inp$ and $\vec{X}_\position$. This would amount to independently dropping the rows of the matrices $\vec{W}_\inp$ and $\vec{W}_\position$. In a typical input to a transformer this corresponds to the input sequence (\emph{e.g.,} a sentence) and is commonly not regularized. The dropout mask is applied after the corresponding input and position embedding is computed. However, in BayesFormer architecture we apply the dropout masks \emph{before} the embedding is computed. This ensures that these embeddings themselves aren't suffering from overfitting and the model is forced to not depend on any single input in the sequence. 

Next, in BayesFormer we modify the application of dropout in the self-attention unit. First, the dropout mask for each layer $i \in \ell$ and each head $j \in k$ is chosen independently.  Second, for a self-attention unit $i \in \ell$ and a head $j \in k$, combining equations~\eqref{eq:multi-headAttention} and \eqref{eq:variationalDistribution} we apply dropout masks $h_\key$, $h_\query$ and $h_\val$ to the vectors $\vec{X}_\key$, $\vec{X}_\query$ and $\vec{X}_\val$ respectively. Note that in a typical self-attention unit the vectors  $\vec{X}_\key$, $\vec{X}_\query$ and $\vec{X}_\val$ are identical yet in BayesFormer we sample \emph{three} independent dropout masks and apply them separately to the vectors. This can be interpreted as projecting the self-attention matrices to \emph{random} low-dimensional subspaces similar to that of efficient transformer architectures such as linformer \cite{wang2020linformer}. Third, we do not apply any other dropout within the self-attention, notably \emph{removing} the dropout that is typically applied before multiplying with $\vec{W}_\val$. This is not surprising, since we have already regularized by applying an independent dropout mask $h_{\val}$.

Finally, in BayesFormer we apply dropout mask to the input of the feed-forward layer following the multi-head self-attention. This input consists of two parts: concatenated output from the multi-head self-attention followed by a layer norm and a skip-connection that is the input to the multi-head self-attention of the previous layer. We apply a dropout mask $h_\mlp$ to this concatenated string before passing it through the feed-forward layer. As in the standard transformer architecture, the dropout application within the feed-forward layer remains unchanged. Figure~\ref{Fig:BayesFormer_summary} summarizes the key changes in BayesFormer compared to a standard transfomer architecture. In this figure, same color represents that the outcomes of random variable is shared, while different colors indicate independence in the outcomes.

Given the BayesFormer architecture, the proof of theorem~\ref{thm:mainTheorem} follows from equations~\eqref{eq:variationalDistribution} and \eqref{eq:multi-headAttention}.

\section{Experiments}
In this section, we summarize the results from the empirical evaluation of the proposed theory. We use four different paradigms that comprehensively illustrates the behavior of the proposed methods. We use language classification tasks, long range sequence understanding tasks, machine translation and active learning. We also explore how this dropout methodology integrates into other efficient transformer architectures (commonly referred to as x-formers). In all the conducted experiments, we control for as many sources of confounding variables as possible between the control and the test arms. This includes non-dropout related hyper-parameters, number of GPUs and/or nodes in distributed training and the seed used in random function invocation during data pre-processing. This helps us understand the contribution of the dropout variable independent of other architectural choices and thus, can be extended to other SOTA architectures when appropriate. Regarding the dropout hyper-parameters themselves, we tune the dropout probability in the small range of $\{0.05, 0.1, 0.2\}$ using a validation set for the task. Where resources are not specified a run of an experiment was performed using a single Nvidia A100 GPU on a single node of a cluster.

\subsection{Pretrain and finetune paradigm for language classification tasks}
\label{subsec:pretrainFinetune}
The most salient application of the transformer architecture in natural language processing is to pretrain large language models (LLMs). LLMs are pretrained on internet scaled corpus using the masked language modelling approach to understand text. They are then used in language classification tasks such as sentiment analysis by finetuning the LLM on the small dataset corresponding to the task. Some common examples of LLMs include BERT~\cite{devlin2018bert}, RoBERTa~\cite{liu2019RoBERTa} which have had significant impact in the last few years.

Typically, finetuning on a target task leads to the potential of over-fitting since the number of labels available are small (compared to pretraining). The ability to perform large number of updates on the target dataset without over-fitting is crucial since it leads to improved performance. However, there is a trade-off between increasing the number of updates and the potential to over-fit and thus, early stopping is commonly employed (\cite{liu2019RoBERTa}).

In this experiment, we pretrain RoBERTa$_{\base}$ and our proposed model BayesFormer on the English wikipedia corpus appended with the Bookcorpus~\cite{zhu2015aligning} totalling a size of 16GB. As described in~\cite{devlin2018bert}, this corpus was used to pretrain the BERT model. For both the models we use the same hyper-parameters as used in~\cite{liu2019RoBERTa}. For BayesFormer, all new dropout units use the dropout probability of 0.1. Both models are pre-trained for 250k steps.

We fine-tune the pretrained models on \emph{eight} target datasets available in the GLUE dataset \cite{wang2018glue}. The tasks are acceptability (CoLA \cite{warstadt2018neural}), 
sentiment classification (SST2 \cite{socher2013recursive}), paraphrase (MRPC \cite{dolan2005automatically}, QQP \cite{qqp}), sentence similarity (STS-B \cite{cer2017semeval}) and natural language inference (MNLI \cite{N18-1101}, QNLI \cite{rajpurkar2016squad}, RTE \cite{bentivogli2009fifth}). For finetuning, we use the same procedure as that of~\cite{liu2019RoBERTa} where we tune the maximum learning rate $\in \{1e-5, 2e-5, 3e-5\}$, the batch size $\in \{16, 32\}$ hyper-parameters for 10 epochs. We use the best checkpoint in the 10 epochs based on the performance on the validation set to evaluate on the dev set. The pretrained models were trained using distributed training on 8 nodes with each node containing 8 Nvidia A100 GPUs. The finetuning runs on the other hand were run on a single Nvidia A100 GPU. We used the fairseq~\cite{ott2019fairseq} code-base to implement the baselines, our method and run the data pre-processing and testing scripts.

\textbf{Results and discussion.} Table~\ref{tab:GLUE} summarizes the results of this experiment. As we can see, BayesFormer performs well on average and improves over RoBERTa$_\base$ on 6 out of the 8 GLUE tasks, comparably on one and performs worse on the other. To further explore the behavior of BayesFormer, we looked into the negative log-likelihood (nll) and the loss evolution during the 10 epochs of fine-tuneing. Figure~\ref{Fig:overfit_nll} show that RoBERTa$_\base$ model very quickly over-fits (starting epoch 3) while BayesFormer does not overfit; both loss and nll continue to decrease with the number of updates. 

\textbf{Further ablations.} We consider further ablation to understand the effect of datasize towards overfitting and thus, the eventual improvement in performance. We pretrain using \emph{only} the English wikipedia dataset for pretraining the LLMs. Thus, the dataset size for pretraining is significantly smaller. We notice that the relative improvement of BayesFormer over RoBERTa$_\base$ in downstream GLUE tasks is higher when pretrained with the smaller English wikipedia corpus. Table~\ref{tab:GLUEAblation} summarizes the results when trained with the smaller dataset. This further supports the fact that BayesFormer is less prone to overfitting compared to that of vanilla transformer.

\begin{table}
\centering
\begin{tabular}{p{2.3cm} P{0.8cm} P{0.8cm} P{0.8cm} P{1.5cm} P{1.5cm} P{1.5cm} P{0.8cm} P{0.8cm} P{0.8cm}}
\hline
 Model & CoLA  & SST2 & MRPC & STSB  & QQP  & MNLI m/mm  & QNLI & RTE  & \textbf{Avg} \\
 & (MCC) & (Acc) & (F1) & (P/S) & (F1/Acc) & (Acc) & (Acc) & (Acc)\\
 \hline \\
 RoBERTa$_\base$ & 62.1 & \textbf{94} & 90.9 & 87.3/87.3  & \textbf{88.5}/91.5 & 84.9/84.9 & 91.4 & 74.7 & 84.6\\
 BayesFormer & \textbf{63.3} & 93.7 & \textbf{92.1} & \textbf{87.7}/\textbf{87.7} & 88.4/91.5 & \textbf{85.3}/\textbf{85.3} & \textbf{91.7} & \textbf{74.8}& \textbf{85} \\
 \hline
\end{tabular}
\caption{\label{tab:GLUE} Dev set results (median of five runs). RoBERTa$_\base$ v/s BayesFormer: pretrained on BERT corpus and finetuned on the GLUE benchmark datasets.}
\end{table}

\subsection{Long-range sequence understanding}

Transformers are useful over prior architectures (e.g., LSTM) for sequential understanding problems since they are able to effectively learn long range contextual dependencies in data effectively. Recently, Long-range arena dataset~\cite{tay2020long} was introduced to provide a standardized evaluation of transformers in the long-context understanding setting. It contains five tasks across domains: Listops~\cite{nangia2018listops}, byte-level text classification~\cite{howard2018universal}, byte-level document retrieval~\cite{guo2016deep}, Image classification on pixels~\cite{krizhevsky2009learning} and Pathfinder\footnote{We report results on the length 14 setting.}~\cite{kim2019disentangling}. The LRA repository provides scripts for data preprocessing, configs for parameters and an \emph{apples-to-apples} comparison environment. We include our proposed approach to the \emph{softmax attention} implementation and use the same configuration values. We set the dropout probability of new dropout operations at 0.05.

\begin{figure}
   \begin{minipage}{0.5\textwidth}
     \centering
     \includegraphics[width=\linewidth]{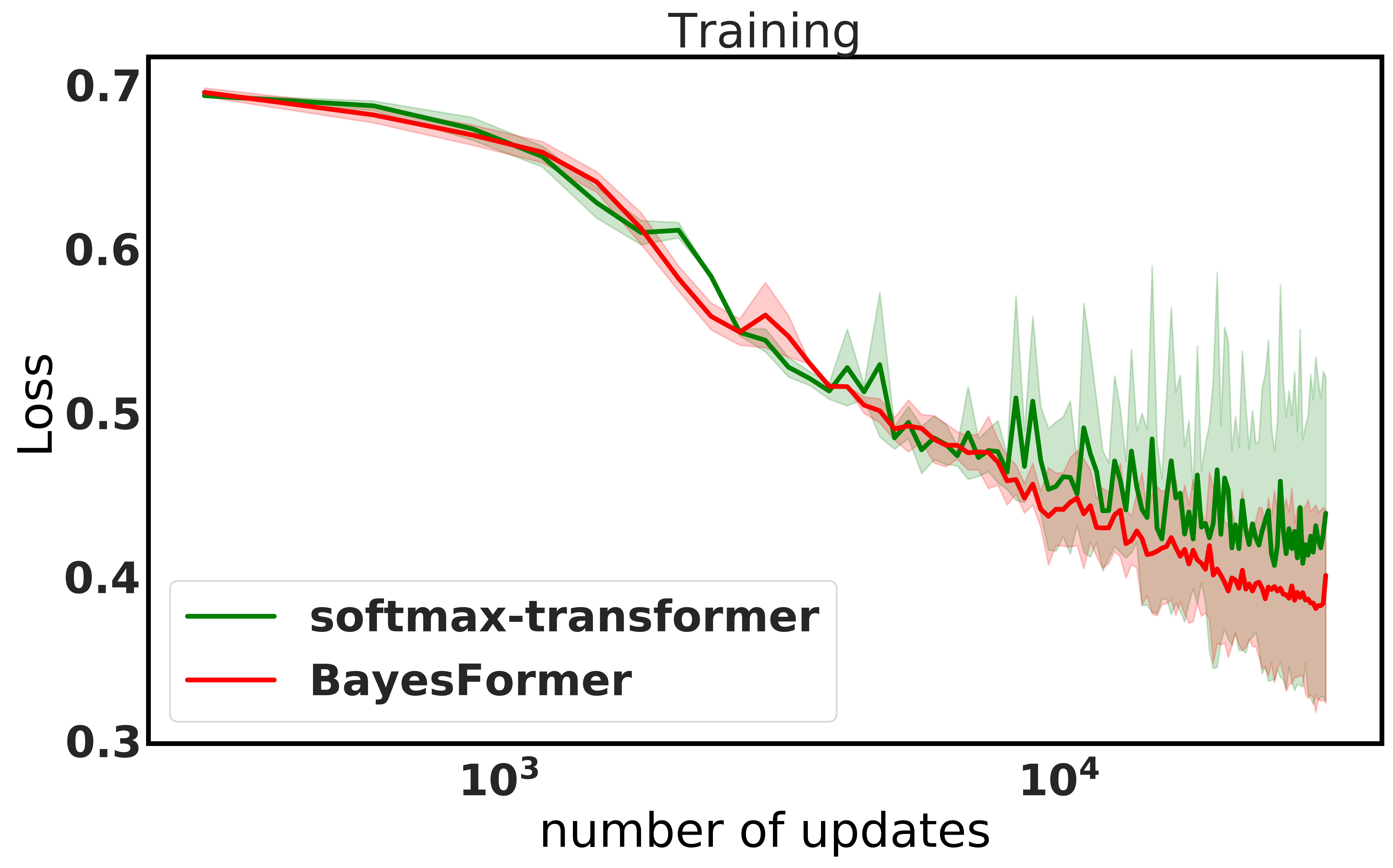}
   \end{minipage}\hfill
   \begin{minipage}{0.5\textwidth}
     \centering
     \includegraphics[width=\linewidth]{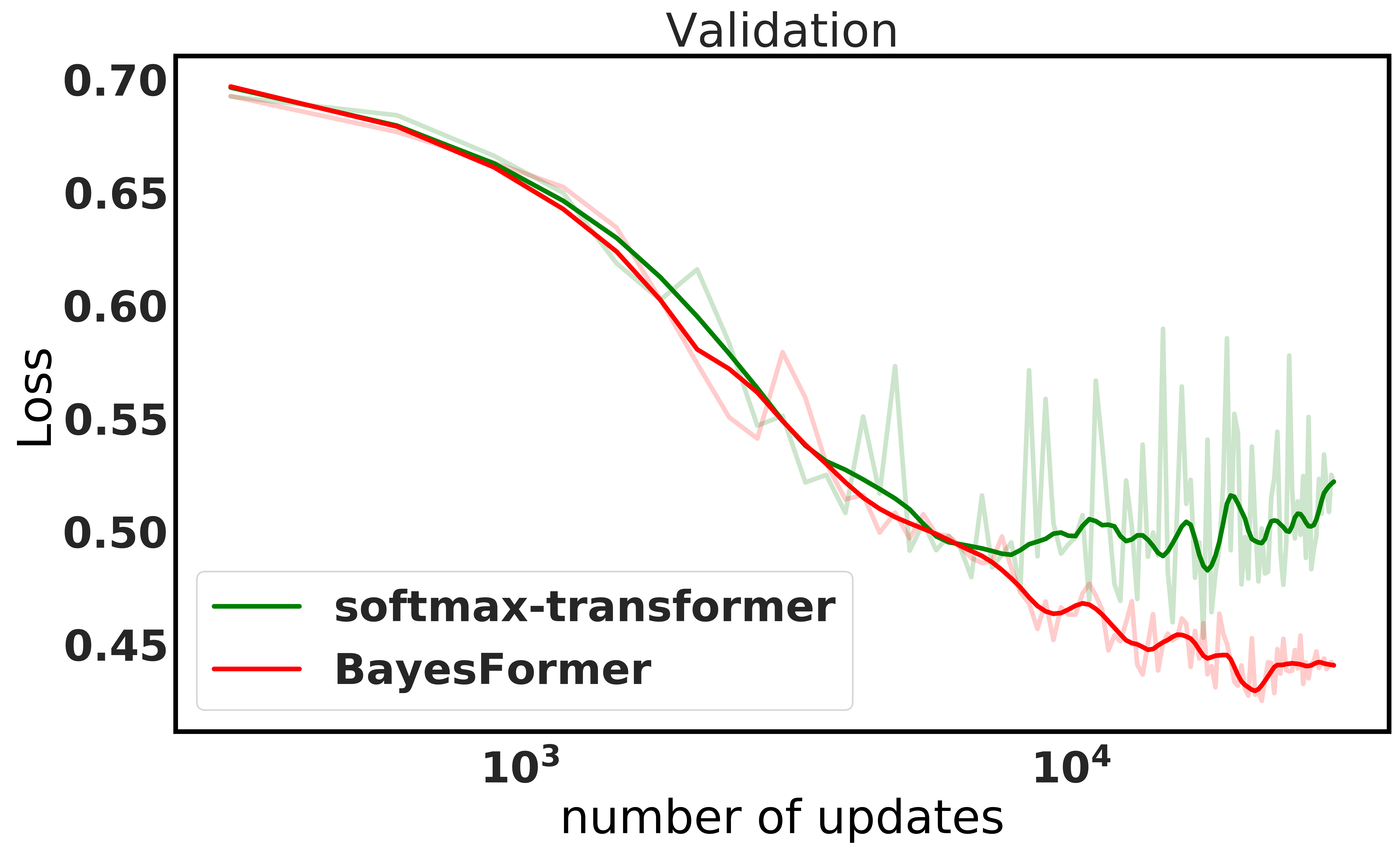}
   \end{minipage}
   \caption{Loss v/s num. of updates on LRA retrieval dataset. Smoothed using lowess.}
   \vspace{0.6cm}
    \label{Fig:overfit_valid_retrieval}
\end{figure}
\begin{table}
\begin{center}
\begin{tabular}{ p{3.5cm} p{1.8cm}}
\hline
 Model & Test accu. \\
 \hline \\
 Linear attention & 17 (0.53) \\
 BayesFormer - linear & \textbf{19.1} (0.48)  \\
 \hline \\
 Linformer & 36.2 (0.17) \\
 BayesFormer - linformer & \textbf{36.6} (0.05) \\
 \hline \\
 Reformer & 18 (0.78) \\
 BayesFormer - reformer & \textbf{18.1} (2.06) \\
 \hline \\
 Performer & 19.4 (4.4) \\
 BayesFormer - performer & \textbf{23.4} (5.4) \\
 \hline \\
 Nystrom attention & 34.6 (8.7) \\
 BayesFormer - Nystrom & \textbf{35.0} (0.75)\\
 \hline
\end{tabular}
\caption{\label{tab:xformer} Bayes dropout on various x-former attentions.}
\end{center}
\end{table}

\textbf{Results and discussion.} Table~\ref{tab:LRA} summarizes the results on the five LRA tasks. The results are median of five different runs. As we can see we consistently improve over the vanilla transformer showcasing the power of BayesFormer. We attribute these improvements to the ability to reduce over-fitting. To validate this, we plot the train and validation losses on the retrieval task as shown in Figure~\ref{Fig:overfit_valid_retrieval}. We see that the softmax attention transformer start to overfit, as seen by the increasing validation loss with number of updates beyond a point, while the loss further continues to decrease in the validation set for BayesFormer. Surprisingly, we find that BayesFormer also is able to fit the training set better than vanilla transformer. This indicates that the new dropout method is also allowing for the transformer to connect contextual information in the input to a longer part of the sequence likely because the dropout forces the network to not rely on any single part of the input. Unlike the regular transformer, BayesFormer needs to depend on a lossy version of the key, query and value vectors while performing self-attention.

\textbf{Further ablations.} We further use the Listops dataset to evaluate the application of the new dropout on xformers to understand if this methodology is compatible with efficient transformer architectures. For each xformer, we use the base attention module that uses the dropout as described in the respective paper and compare against the dropout application as described by the above theory \emph{before} and \emph{after} the attention units. In particular, we do not modify the way dropout is applied within the attention units and treat all operations within the attention unit as a \emph{black-box}. We consider linear attention~\cite{katharopoulos2020transformers}, linformer~\cite{wang2020linformer}, reformer~\cite{kitaev2020reformer}, performer~\cite{choromanski2020rethinking} and Nystrom attention~\cite{xiong2021nystromformer} variants for the transformer architecture. All experiments were run using the code released by the Nystromformer paper authors.

Initial evidence indicates that the theoretically grounded version of dropout proposed in this paper can indeed improve architectures beyond those it was derived for. Thus, we recommend future SOTA models to test this methodology of dropout to improve the performance.
Table~\ref{tab:xformer} summarizes the results which is a median taken over 5 random runs.

\begin{table}
\begin{center}
\begin{tabular}{ p{2cm} P{1.5cm} P{1.5cm} P{1.5cm} P{1.5cm} P{2cm} P{0.8cm} }
\hline
 Model & Listops & Image & Text & Retrieval & Pathfinder & \textbf{Avg.} \\
 \hline \\
 Transformer & 36 (0.3) & 38.6 (0.6) & 64.8 (0.4) & 78.4 (0.7)  & 69.8 (0.5) & 57.5\\
 BayesFormer & \textbf{36.1} (0.4) & \textbf{39.3} (0.6) & \textbf{65.7} (0.2) & \textbf{80.8} (0.1)& \textbf{72.8} (0.5) & \textbf{58.9} \\
 \hline
\end{tabular}
\caption{\label{tab:LRA} Test set results (median of five runs) on LRA tasks.}
\end{center}
\end{table}

\subsection{Machine Translation}

Machine translation is another important task in Natural language processing where the goal is to translate sentences in a source language to a sentence in the target language that preserves the meaning. This can naturally be modelled as a sequence-to-sequence task and transformer is typically used to achive SOTA results. To evaluate BayesFormer, we consider the IWSLT '14 German to English translation task available in the fairseq repository~\cite{ott2019fairseq}. We use the same scripts for preprocessing the data into train, validation and test splits. The dataset contains 160239 train examples, 7283 validation examples and 6750 test sentences.

\begin{table}
\begin{center}
\begin{tabular}{ p{2.2cm} p{1.5cm} p{1.7cm}}
\hline
 Model & Val BLEU & Test BLEU \\
 \hline \\
 Transformer & 35 (0.1) & 34.5 (0.1) \\
 BayesFormer & \textbf{35.5} (0.1) & \textbf{34.8} (0.1) \\
 \hline
\end{tabular}
\caption{\label{tab:MTL} IWSLT '14 de $\rightarrow$ en test BLEU scores}
\end{center}
\end{table}

We use the architecture \texttt{transformer\_iwslt\_de\_en} available in fairseq repository as the baseline model and apply the proposed dropout methodology to this model. The number of parameters in this model is 34.5M. We do not modify any hyper-parameter apart from the dropout probability on the newly added dropout module which is set to 0.05. Thus, we do not change the dropout probabilities of those that are common to both the baseline and BayesFormer. Both the models are trained for 500k steps and the best checkpoint, as measured by the validation set BLEU score, is used to obtain BLEU scores on the test set. We apply label smoothing and use the \texttt{LabelSmoothedCrossEntropyCriterion} to train the model.

\textbf{Results.} Table~\ref{tab:MTL} summarizes the BLEU score~\cite{papineni2002bleu} on both the validation set and the test set. BayesFormer achieves a higher validation BLEU and consequently a  higher test BLEU score over the baseline transformer variant.

\subsection{Active learning}
Active learning aims to smartly label a limited pool of samples that improves the classifier performance. Labelled data is typically costly to acquire and thus, active learning methods help with cutting down total labelled data needed for comparable performance~\cite{settles2009active}. Here we consider the \emph{pool-based} setup where we can access any subset of data to obtain labels. Typical works (\emph{e.g.,} ~\cite{siddhant2018deep,dor2020active}) in active learning consider the multi-round setting, where the active learning algorithm can obtain labels for a partial subset of data, retrain the classifiers and use that to acquire more data. However, these schemes are highly impractical for real-world systems. In most real-world systems, retraining classifiers is as costly, if not more, as obtaining the labels themselves. Hence, here we instead consider the harder single-round protocol of active learning where the sampling algorithm assigns \emph{scores} to all the unlabelled data at once, and then labels are collected for the \emph{top-k} data points based on this score.

\begin{figure}
  \begin{center}
    \includegraphics[width=\textwidth]{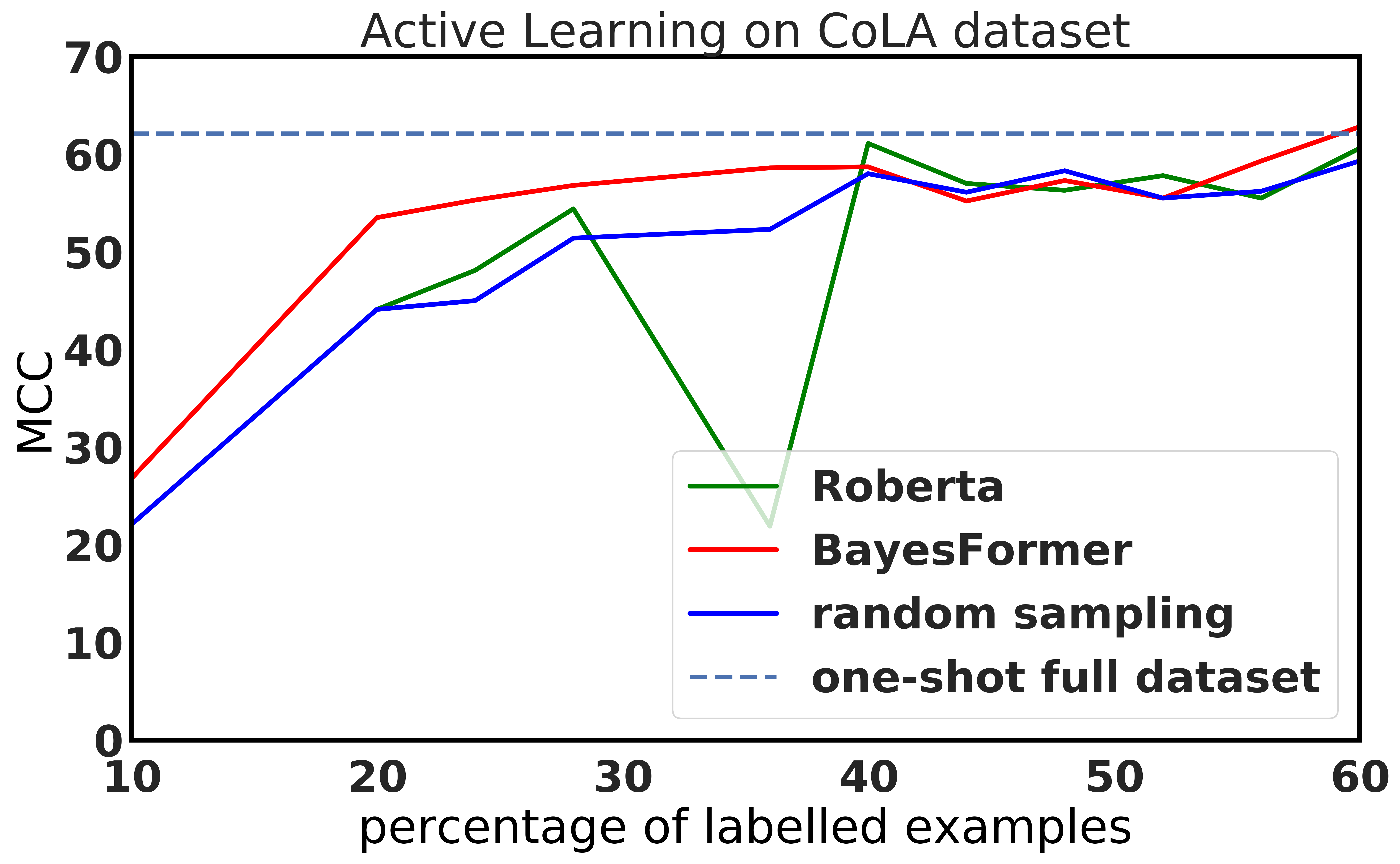}
  \end{center}
  \caption{Active sampling during finetuning on CoLA dataset starting from the same pretrained checkpoint.}
  \label{fig:activeLearning}
\end{figure}

To test the effectiveness of the uncertainty estimates produced by the BayesFormer, we use the MCDropout~\cite{gal2016dropout} based BALD~\cite{gal2017deep,houlsby2011bayesian} acquisition function (called MC-BALD) and compare it against two baselines: MC-BALD applied to the vanilla transformer model and random-sampling. The same pretrainined checkpoint of RoBERTa model is used as a starting point and active sampling is used to select examples for finetuning. We reuse the CoLA dataset from subsection~\ref{subsec:pretrainFinetune}. We warmstart by first finetuning the checkpoint on 10\% of randomly chosen sentences. To obtain the MC-BALD scores for BayesFormer and RoBERTa, we perform 11 forward passes on each point. The remaining 90\% of the CoLA dataset is scored using the three different strategies (\emph{i.e.,} MC-BALD for BayesFormer, RoBERTa and random score). We then finetune by choosing top-k based on these scores, for various values of $k$. Figure~\ref{fig:activeLearning} shows the dev set MCC of the models, based on the percentage of labelled data used. We see that when the labelling budget is low, BayesFormer has a significantly higher MCC (by > 4 points) compared to RoBERTa and random sampling. And as the budget increases, the accuracy of all the three methods catch up as expected. This indicates that the estimates obtained by BayesFormer are more reliable compared to vanilla MCDropout application. A surprising side-effect of this experiment was that sometimes its better to \emph{throw away} some labels to \emph{improve} the dev set accuracy. Indeed, the best MCC for BayesFormer was 62.8 which outperformed the MCC of 59.6 obtained when trained on all data.

\begin{table}
\begin{center}
\begin{tabular}{ p{2.5cm} P{1cm} P{1cm} P{1cm} P{2cm} P{2cm} P{1cm} P{1cm}}
\hline
 Model & CoLA (MCC) & SST2 (Acc) & MRPC (F1) & STSB (P/S)  & MNLI m/mm (Acc) & QNLI (Acc) & RTE (Acc) \\
 \hline \\
 RoBERTa$_\base$ & 21.3 & 91.3 & 89.5 & 84.45/84.22  & 82.1/82.1 & 89.9 & 59.2 \\
 BayesFormer & \textbf{32.1} & \textbf{91.5} & \textbf{90.9} & \textbf{85.58}/\textbf{85.26} & \textbf{83.3}/\textbf{83.3} & \textbf{91.3} & \textbf{65} \\
 \hline
\end{tabular}
\caption{\label{tab:GLUEAblation} Dev set results (median of five runs). RoBERTa$_\base$ v/s BayesFormer: pretrained on English wikipedia corpus (small) and finetuned on the GLUE benchmark datasets.}
\end{center}
\end{table}

\section{Conclusion}

In this paper, we extend the approximate variational inference lens of dropout to Transformer architectures. This helped us derive a new dropout structure with mathematical grounding. We verified the theory empirically on a wide variety of tasks. Based on the results of this paper, we suggest a few future directions. First, it would be interesting to understand the \emph{interplay} of this dropout method with other techniques used to reduce over-fitting. In this paper, we briefly allude to it through ablation on dataset size, but we believe more work is needed. Second, we are also excited to understand its effect on other applications of large transformer models such as zero/few-shot learning and text generation. Third, we would like to  understand model robustness and brittleness to adversarial examples for BayesFormer.

\section{Acknowledgements}

KAS is grateful to Darsh Shah for helping setup the fairseq environment.
\bibliographystyle{plainnat}
\bibliography{refs}

\appendix

\section{Appendix}

\section{Extended Related Work}

Our work falls at the intersection of Sequential decision making and Transformers. As such these areas of study are extensively explored independently.

\textbf{Uncertainty estimates in sequential decision making.} Uncertainty quantification can be primarily motivated in the context of sequential decision making, and in particular, the field of multi-armed bandits~\cite{slivkins2019introduction}. As illustrated in~\cite{gal2016dropout}, neural networks with uncertainty estimates can be used to perform reinforcement learning~\cite{sutton2018reinforcement}. They explore the Thompson sampling algorithm~\cite{russo2018tutorial} and show that it outperforms the popular $\epsilon$-greedy algorithm~\cite{langford2007epoch}. Multi-armed bandits are the primary choice of techniques used for exploration in recommender systems~\cite{langford2007epoch,li2010contextual} due to its simplicity and effectiveness. Beyond recommender systems uncertainty based exploration is used in other applications such as advertising~\cite{avadhanula2021stochastic} and ride-share platforms~\cite{han2021budget}. Beyond multi-armed bandits, uncertainty estimates are also used for reinforcement learning algorithms such as~\cite{osband2016deep,lowrey2018plan}.

The other salient application of uncertainty in sequential decision making is in active learning~\cite{settles2009active}. Many classical active learning algorithms such as uncertainty sampling~\cite{lewis1994heterogeneous}, Query-by-committee~\cite{melville2004diverse} and more recent ensemble approaches such as MCDropout~\cite{gal2016dropout}, BALD~\cite{gal2017deep}, Bayes-by-backprop~\cite{blundell2015weight} and MAAL~\cite{ebrahimi2020minimax} all rely on either implicit or explicit uncertainty quantification of predictions of the current model to acquire the next batch of data.

\textbf{Transformers.} Following the initial breakthrough work of \cite{vaswani2017attention}, Tranformer architectures have received a ton of attention. Some directions include efficient transformers~\cite{tay2020efficient} where the goal is to reduce the complexity of the costly self-attention computation while keeping the performance near parity. The survey above covers most of the innovations in this direction including benchmarks, models and applications. On the application side, transformer architecture has been used in computer vision~\cite{han2022survey}, graphs~\cite{min2022transformer} and natural language processing~\cite{lin2021survey}. Recently, transformer architectures have also started to be deployed in practice~\cite{sun2019bert4rec,pei2019personalized}. Tangentially related to our work is that of \cite{wu2021unidrop} where they explore many different positions dropout could be applied in a transformer architecture. Similar to our work, they notice the presence of over-fitting. However, different from our work, none of their dropout methods include the ones proposed in this paper. Another work that on the surface is related to our work is \cite{xue2021bayesian}. In this work, the authors propose to use MCDropout naively on \emph{existing} application of dropout in the context of speech recognition. First, as we show in this paper for us to have formal guarantees on the posterior, we would need to modify the way dropout is applied. In that respect, the prior work is not truly \emph{Bayesian}. Second, the goals of the two papers are different. Indeed, they are interested in improving speech recognition benchmarks, while we are interested in principled understanding irrespective of the application domain.

\section{Reproducibility details for experiments}

In this section we specify how we chose the hyper-parameters for the baseline as well as BayesFormer in the various experiments.

\textbf{Pretraining}. For the pretraining of RoBERTa$_\base$ model, we used the same settings from~\cite{liu2019RoBERTa}. For BayesFormer, we also chose the same set of hyper-paramters and settings. We did not do any hyper-paramter tuning at this step.

\textbf{Finetuning.} For finetuning on the various GLUE tasks, for both RoBERTa$_\base$ and Bayesformer, we tuned \emph{three} hyper-parameters. Learning rate $ \in \{ 1e-5, 2e-5, 3e-5, 4e-5, 5e-5, 1e-4\}$, batch size in $\in \{ 16, 32\}$ and the dropout probability $\in \{0.05, 0.1, 0.2\}$. We choose the best model based on the validation set metric.

\textbf{LRA tasks.} For LRA tasks, we did not change any hyper-parameter or settings. We cloned the code made available by authors of \cite{xiong2021nystromformer} and implemented our algorithm in it. We used the same settings as softmax-transformer for BayesFormer. For the ablation with xformers, we copied the corresponding settings from the respective xformer.

\textbf{Machine Translation.} For the machine translation task, we once again did not tune any hyper-parameter. We directly leveraged the scripts available as part of fairseq and ran it using the standard settings.

\textbf{Active learning.} For active learning, we picked the best model for CoLA dataset from above for both RoBERTa$_\base$ and Bayesformer and ran our experiments using that setting.
\end{document}